\documentclass[10pt,twocolumn,letterpaper]{article}

\usepackage{cvpr}              % To produce the CAMERA-READY version
\usepackage{amssymb}
\usepackage[dvipsnames]{xcolor}
\usepackage{tabularx}
\usepackage{multirow} % used by tables
\usepackage{makecell} % used by tables
\usepackage{colortbl} % used by tables
\usepackage[Export]{adjustbox} % provides valign of images and align
\usepackage{arydshln} % dashed lines in tables
\usepackage{colortbl}
\usepackage{pifont} % for checkmark symbols
\usepackage{tikz} % frame around image
\usepackage{bm}
\usepackage{caption}
\usepackage{subcaption} % subtable
\def\eg{\emph{e.g.}\xspace} 
\def\ie{\emph{i.e.}\xspace} 

\def\wrt{\emph{w.r.t.}\xspace}

\newcommand{\ours}{\textit{GaussFusion}\xspace}

\colorlet{colorFst}{Green!25}       % first
\colorlet{colorSnd}{SpringGreen!45} % second
\colorlet{colorTrd}{Yellow!30}      % third
\colorlet{colorLow}{darkgray!30}    % low-light color
\newcommand{\fs}{\cellcolor{colorFst}\bf}   % first
\newcommand{\nd}{\cellcolor{colorSnd}}      % second
\newcommand{\rd}{\cellcolor{colorTrd}}      % third

\newcolumntype{R}{>{\raggedleft\arraybackslash}X}

\usepackage{tocloft}

\definecolor{cvprblue}{rgb}{0.21,0.49,0.74}
\usepackage[pagebackref,breaklinks,colorlinks,allcolors=cvprblue]{hyperref}

\def\paperID{***} % *** Enter the Paper ID here
\def\confName{CVPR}
\def\confYear{2026}

\title{GaussFusion: Improving 3D Reconstruction in the Wild with \\A Geometry-Informed Video Generator}

\author{Liyuan~Zhu\textsuperscript{1} \quad
Manjunath Narayana\textsuperscript{2} \quad
Michal Stary\textsuperscript{2} \quad
Will Hutchcroft\textsuperscript{2} \quad \\
Gordon Wetzstein\textsuperscript{1} \quad
Iro Armeni\textsuperscript{1} 
\quad
\vspace{5px}
\\
{ \textsuperscript{1}{Stanford University} \quad 
\textsuperscript{2}{Zillow Group}
} \\
}

\vspace{-5px}

\begin{document}

\twocolumn[{%
	\renewcommand\twocolumn[1][]{#1}%
	% \vspace{-10pt}
        \maketitle
	\begin{center}
        %\vspace{0mm}
    \begin{tabular}{ccccc}
    % Image spanning all columns
    \multicolumn{5}{c}{%
        \includegraphics[width=1.\linewidth]{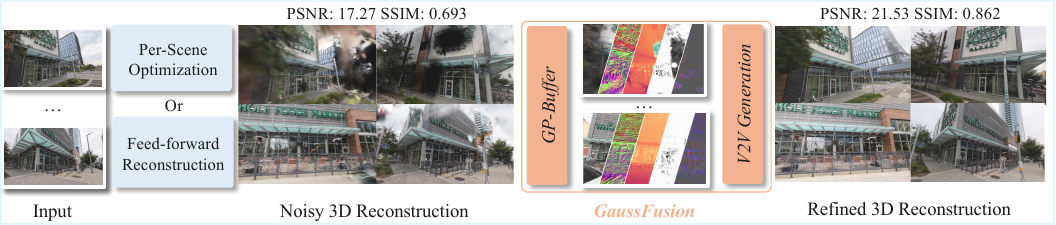}%
    }
\end{tabular}
	\captionof{figure}{
\textbf{\ours Overview.} Given multi-view images as input, we first obtain an initial 3D Gaussian splatting (3DGS)~\cite{kerbl3Dgaussians} reconstruction using either per-scene optimization or a feed-forward reconstruction  model~\cite{xu2024depthsplat,jiang2025anysplat}. The reconstruction often exhibits visual noise and geometric artifacts (PSNR = 17.27, SSIM = 0.693). Our method \textit{GaussFusion}, first rasterizes a \textit{Gaussian Primitives Buffer (GP-Buffer)} encoding color, depth, normals, opacity, and covariance, and then employs a novel GP-Buffer-to-RGB video generation framework to obtain clean, artifact-free novel-view renderings and enhance the underlying 3D representation. The resulting reconstruction achieves significantly improved rendering quality and consistency (PSNR = 21.53, SSIM = 0.862).
}

	\label{fig:teaser}
	% \vspace{3mm}
	\end{center}    
}]

\maketitle
\begin{abstract}
We present \ours, a novel approach for improving 3D Gaussian splatting (3DGS) reconstructions in the wild through geometry-informed video generation. \ours mitigates common 3DGS artifacts, including floaters, flickering, and blur caused by camera pose errors, incomplete coverage, and noisy geometry initialization. Unlike prior RGB-based approaches limited to a single reconstruction pipeline, our method introduces a geometry-informed video-to-video generator that refines 3DGS renderings across both optimization-based and feed-forward methods. Given an existing reconstruction, we render a Gaussian primitives video buffer encoding depth, normals, opacity, and covariance, which the generator refines to produce temporally coherent, artifact-free frames. We further introduce an artifact synthesis pipeline that simulates diverse degradation patterns, ensuring robustness and generalization. \ours achieves state-of-the-art performance on novel view synthesis benchmarks, and an efficient variant runs in real time at 16 FPS while maintaining similar performance, enabling interactive 3D applications.  [\href{http://research.zhuliyuan.net/projects/GaussFusion/}{Project Page}]

\end{abstract}    
\section{Introduction}
\label{sec:intro}
Photorealistic 3D reconstruction and novel-view synthesis are fundamental problems in computer vision, with applications in virtual reality, autonomous driving, and robotics. 3D Gaussian splatting~\cite{kerbl3Dgaussians} has emerged as a popular representation for high-quality reconstruction and rendering.
More recently, feed-forward methods~\cite{wang2024dust3r,wang2025vggt,zhang2025flare,keetha2025mapanything} predict 3D Gaussian parameters directly in a single network pass~\cite{smart2024splatt3r,zhang2024gslrm,chen2024mvsplat,charatan2024pixelsplat,chen2024mvsplat360,xu2024depthsplat,jiang2025anysplat,ziwen2025longlrm}, enabling much faster and more robust reconstructions compared to traditional optimization-based pipelines. 
However, despite these advances, current methods still suffer from artifacts in sparse-view and under-captured scenarios, and degrade significantly at novel views far from training views (\cref{fig:teaser}, noisy).

To address these limitations, several methods~\cite{chan2023genvs,wu2023reconfusion,yu2024viewcrafter,liu2024dgsenhancer,liu2024reconx,zhong2025guidevd,teng2025fvgen,ren2025gen3c,wang2025videoscene} have explored leveraging generative priors to enhance 3D reconstruction by generating dense novel-view images. Recent approaches such as Difix3D~\cite{wu2025difix3d}, GenFusion~\cite{Wu2025GenFusion}, and ExploreGS~\cite{Kim_2025_ExploreGS} perform post-hoc artifact correction using generative models conditioned only on RGB(D) renderings. 
In practice, this limited conditioning makes these methods effective primarily for correcting minor blurring or color shifts, but they fail to address more challenging artifacts—such as large floaters, missing regions, and geometric errors—where reliance on color alone leads to significant ambiguity. Further, we find that approaches trained for optimization-based 3DGS pipelines fail to generalize to feed-forward reconstructions, which exhibit distinct degradation patterns, often introducing additional artifacts and reduced reconstruction quality (see \cref{tab:feedforward_refine} and \cref{fig:image_refine}). Similarly, MVSplat360~\cite{chen2024mvsplat360} refines feed-forward reconstructions but fails to generalize to optimization-based pipelines, as it is tightly coupled to a specific feed-forward model~\cite{chen2024mvsplat}. We argue that a truly practical refinement model should be agnostic to the reconstruction paradigms, as real-world 3D data may originate from different sources: optimization-based~\cite{kerbl3Dgaussians,zhu2025_loopsplat}, feed-forward pipelines~\cite{ye2024no,xu2024depthsplat}, or even 3D generation~\cite{szymanowicz2025bolt3d,WorldLabs2025biggerbetter}.

This raises a key question: 
\textit{How can we train a single high-quality reconstruction refinement model that generalizes across different 3DGS paradigms?}
Our key insights are twofold: \emph{(i)} existing methods leverage only partial information (color) from Gaussian splats, whereas incorporating full 3DGS primitives—such as depth, opacity, normals, and covariances—provides richer cues for identifying and correcting artifacts regardless of the source. Thus a video-to-video generative model can encode informative cues about splat artifacts and better refine novel views; and 
\emph{(ii)} a comprehensive artifact simulation strategy that exposes the model to diverse degradation patterns as found in multiple methods and datasets during training is crucial for generalization. 

We present \ours, a video-to-video generative model for robust 3D reconstruction that features as key component the \textit{GP-Buffer}, a pixel-aligned video representation that encodes multi-modal cues from the 3D reconstruction, including appearance and geometric information. A \textit{Geometry Adapter} module further injects these appearance and geometry features into the transformer backbone of the video generator, enabling geometry-aware conditioning. 
\ours translates low-quality 3D renders into photorealistic, artifact-free renders (see \cref{fig:teaser}, refined) with a high efficiency (16 FPS).

Our main contributions are as follows:
\begin{itemize}
    \item A geometry-informed video-to-video generation model, \ours, conditioned on 3DGS geometric renders, effective for artifact removal across diverse reconstruction pipelines. 
    \item A comprehensive artifact simulation strategy that synthesizes a novel dataset of 75K+ videos including a wide range of realistic reconstruction degradations, enabling robust and generalizable refinement across both optimization and feed-forward settings.
    \item A novel finetuning recipe for a few-step variant, enabling efficient on-the-fly refinement during rendering and real-time applications.
\end{itemize}

\section{Related Work}
\paragraph{3D Reconstruction and Novel-view Synthesis.} 3D Gaussian splatting~\cite{kerbl3Dgaussians} has emerged as the dominant representation for 3D reconstruction, offering explicit scene modeling and real-time rendering efficiency. Several works~\cite{Niemeyer2021Regnerf,kangle2021dsnerf,turkulainen2024dnsplatter,zhang2024fregs,hyung2024effective} improve the stability of optimization-based 3DGS through additional regularizations on geometry or appearance, while others~\cite{niemeyer2025radsplat,kotovenko2025edgs,Foroutan2024arxiv,kheradmand2024_mcmc} enhance initialization and densification strategies for higher fidelity and robustness. More recently, feed-forward methods~\cite{wang2024dust3r,wang2025vggt,zhang2025flare,keetha2025mapanything} have replaced iterative optimization with direct prediction of Gaussian parameters~\cite{smart2024splatt3r,chen2024mvsplat,charatan2024pixelsplat,xu2024depthsplat,jiang2025anysplat,ye2024no}, achieving faster and more robust reconstruction. % from images. 

Overall, both paradigms offer complementary strengths for producing 3DGS representations, yet no existing method can be applied across both paradigms to address hard-to-prevent artifacts.

\paragraph{Generative Priors for 3D Reconstruction.}
Diffusion models have been introduced as powerful priors for 3D reconstruction~\cite{chan2023genvs,wu2023reconfusion,yu2024viewcrafter,liu2024reconx,zhou2025stable,ren2025gen3c,jiang2025geo4d,teng2025fvgen,bao2025free,zhong2025guidevd,paliwal2025ri3d,lu2025matrix3d}, improving view interpolation and completion through generative regularization. Building on this idea, more recent approaches apply diffusion directly to refine 3DGS rendering. 3DGS-Enhancer~\cite{liu2024dgsenhancer} and ExploreGS~\cite{Kim_2025_ExploreGS} train a video diffusion model to refine 3DGS renderings with novel camera trajectory sampling strategies. MVSplat360~\cite{chen2024mvsplat360} trains a video diffusion model to directly refine the 3DGS rendering from a feed-forward reconstruction model~\cite{chen2024mvsplat}. Difix3D+~\cite{wu2025difix3d} finetunes a singe-step image diffusion model that cleans up single-view renders with artifacts. 
FlowR~\cite{fischer2025flowr} proposes a multi-view flow matching model that directly flows from corrupt views to cleaned views. Concurrent works such as FixingGS~\cite{wang2025fixinggs} and GSFixer~\cite{yin2025gsfixer} leverage score distillation~\cite{poole2022dreamfusion} or foundation models~\cite{wang2025vggt,caron2021emerging} as conditioning to refine artifacts. 
Despite progress, existing video-based solutions rely on suboptimal conditioning strategies that underutilize geometric cues inherent in the reconstruction itself and remain very slow at inference. 
Moreover, they often lack comprehensive artifact simulation pipelines, limiting generalization across reconstruction paradigms. 
We adopt an efficient video-to-video model that fully exploits geometric cues and generalizes to reconstructions from different sources.

\paragraph{Video-to-video Translation.} 
Video translation has evolved from GAN-based approaches such as vid2vid~\cite{wang2018vid2vid} and Mocycle-GAN~\cite{mocyclegan2019} to modern diffusion- and flow-based generators. Approaches like Rerender-A-Video~\cite{yang2023rerender}, TokenFlow~\cite{geyer2023tokenflow}, FlowVid~\cite{liang2023flowvid}, and StreamV2V~\cite{liang2024streamv2v} improve temporal consistency and extend video editing to longer sequences. More recently, VACE~\cite{vace_wan} unifies diverse video-to-video translation tasks under a single framework. MotionStream~\cite{shin2025motionstream} enables real-time, interactive video-to-video generation by adapting diffusion distillation methods~\cite{yin2024dmd,huang2025selfforcing}.

However, these methods are primarily designed for appearance manipulation or text-conditioned video synthesis, not for refining geometrically grounded 3D renderings. 
In contrast, we repurpose the T2V model Wan~\cite{wan2025} for efficient video-to-video refinement, enabling geometry-aware enhancement of 3D Gaussian splatting.
 
\section{GaussFusion}
Consider an existing 3DGS reconstruction represented by a set of 3D Gaussian primitives $\mathcal{G}$ and its corresponding input images $\{\mathbf{I}^v\in \mathbb{R}^{H \times W \times 3}\}_{v=1}^{V}$ with known camera parameters $\{(\mathbf{T}^v\in SE(3), \mathbf{K}^v\in \mathbb{R}^{3\times3})\}_{v=1}^{V}$, $V$ being the number of input views. Our \textbf{goal} is to enhance novel-view renderings and use them to further optimize $\mathcal{G}$. 

\subsection{Preliminaries}
\paragraph{3D Gaussian Splatting.}
3DGS~\cite{kerbl3Dgaussians} represents a scene as a collection of $N$ anisotropic 3D Gaussian primitives, $
\mathcal{G} = \{(\mathbf{p}_i, \alpha_i, \mathbf{c}_i, \mathbf{q}_i, \mathbf{s}_i) \}_{i=1}^{N},
$: position (mean) $\mathbf{p}_i$, opacity $\alpha_i$, Spherical Harmonics (SH) coefficients $\mathbf{c}_i$, and a 3D covariance matrix $\mathbf{\Sigma}_i$ factorized into a scaling vector $\mathbf{s}_i \in \mathbb{R}^3$ and a rotation quaternion $\mathbf{q}_i \in \mathbb{R}^4$, such that $\mathbf{\Sigma}_i = \mathbf{R}_i \mathbf{S}_i \mathbf{S}_i^\top \mathbf{R}_i^\top$. For view synthesis, given a view transformation $\mathbf{W}$ and the Jacobian $\mathbf{J}$ of the affine approximation of the perspective projection, the 3D covariance $\mathbf{\Sigma}_i$ is projected into a 2D covariance $\mathbf{\Sigma}_i'$:
    $\mathbf{\Sigma}_i' = \mathbf{J} \mathbf{W} \mathbf{\Sigma}_i \mathbf{W}^\top \mathbf{J}^\top$.
The final color $\mathbf{C}$ for a pixel $\mathbf{u}$ is computed by $\alpha$-blending all $N$ Gaussians, sorted by depth, that overlap the pixel:
\begin{equation}
    \mathbf{C}(\mathbf{u}) = \sum_{i=1}^{N} \mathbf{c}_i' \gamma_i \prod_{j=1}^{i-1} (1 - \gamma_j)
    \label{eq:alphablending}
\end{equation}
where $\mathbf{c}_i'$ is the view-dependent color evaluated from the SH coefficients $\mathbf{c}_i$. The contribution $\gamma_i$ is the product of the learned opacity $\alpha_i$ and the 2D Gaussian function evaluated at the pixel center $\mathbf{u}$ with projected mean $\mathbf{\mu}_i'$.

The Gaussian parameters are initialized from sparse structure-from-motion point clouds~\cite{schoenberger2016colmap} and then optimized via stochastic gradient descent to minimize a loss function between the rendered $\mathbf{C}$ and the ground-truth $\mathbf{C}_{\text{gt}}$ images. It is typically a combination of an $L_1$ and a D-SSIM term:
\begin{equation}
    \mathcal{L} = (1 - \lambda) \mathcal{L}_{L_1}(\mathbf{C}, \mathbf{C}_{\text{gt}}) + \lambda \mathcal{L}_{\text{D-SSIM}}(\mathbf{C}, \mathbf{C}_{\text{gt}}).
    \label{eq:loss}
\end{equation}

\paragraph{Feed-Forward 3DGS Reconstruction Models} learn to directly predict a complete set of 3D Gaussian parameters from a small set of posed/unposed input images 
% $\{(I^v, \mathbf{T}^v, \mathbf{K}^v)\}_{v=1}^{V}$
~\cite{zhang2024gslrm,ye2024no,charatan2024pixelsplat,xu2024depthsplat}. A neural network $f_\theta$ aggregates geometric and appearance cues across views to output all parameters for each Gaussian (position $\mathbf{p}_j$, rotation $\mathbf{q}_j$, scale $\mathbf{s}_j$, opacity $\alpha_j$, and SH color $\mathbf{c}_j$) in a single forward pass. Feed-forward approaches eliminate the need for per-scene optimization, and are typically supervised by 3DGS rendering loss (Eq.~\ref{eq:loss}).

\paragraph{Flow-based Video Generation.}
Flow matching~\cite{liu2022flow} provides a continuous-time framework for learning generative processes, forming the basis of modern generation models~\cite{wan2025,esser2024scaling}. 
Given a target sample $x_1$ (\eg, image or video), random noise $x_0 \sim \mathcal{N}(0,I)$, and a timestep $t \in [0,1]$, the intermediate latent $x_t$ is defined by:
\begin{equation}
    x_t = t x_1 + (1-t)x_0,
\end{equation}
whose ground-truth velocity is
$
    v_t = \frac{dx_t}{dt} = x_1 - x_0.
$
The flow network $u_\theta(x_t, c, t)$ is trained to predict $v_t$ using a mean squared error objective:
\begin{equation}
    \mathcal{L} = \mathbb{E}_{x_0, x_1, c, t} \left[\|u_\theta(x_t, c, t) - v_t\|^2\right],
\label{eq:fm}
\end{equation}
where $c$ denotes conditioning signals (\eg, text, images, or videos). 
Videos are encoded into a latent space via a spatio-temporal autoencoder, and flow matching operates entirely therein. At inference time, the learned velocity field defines an ordinary differential equation
    $\frac{dx_t}{dt} = u_\theta(x_t, c, t),$
integrated from $t{=}0$ to $1$ to map Gaussian noise $x_0$ into a generated sample $x_1$.

\begin{figure*}[t]
    \centering
    \includegraphics[width=.88\linewidth]{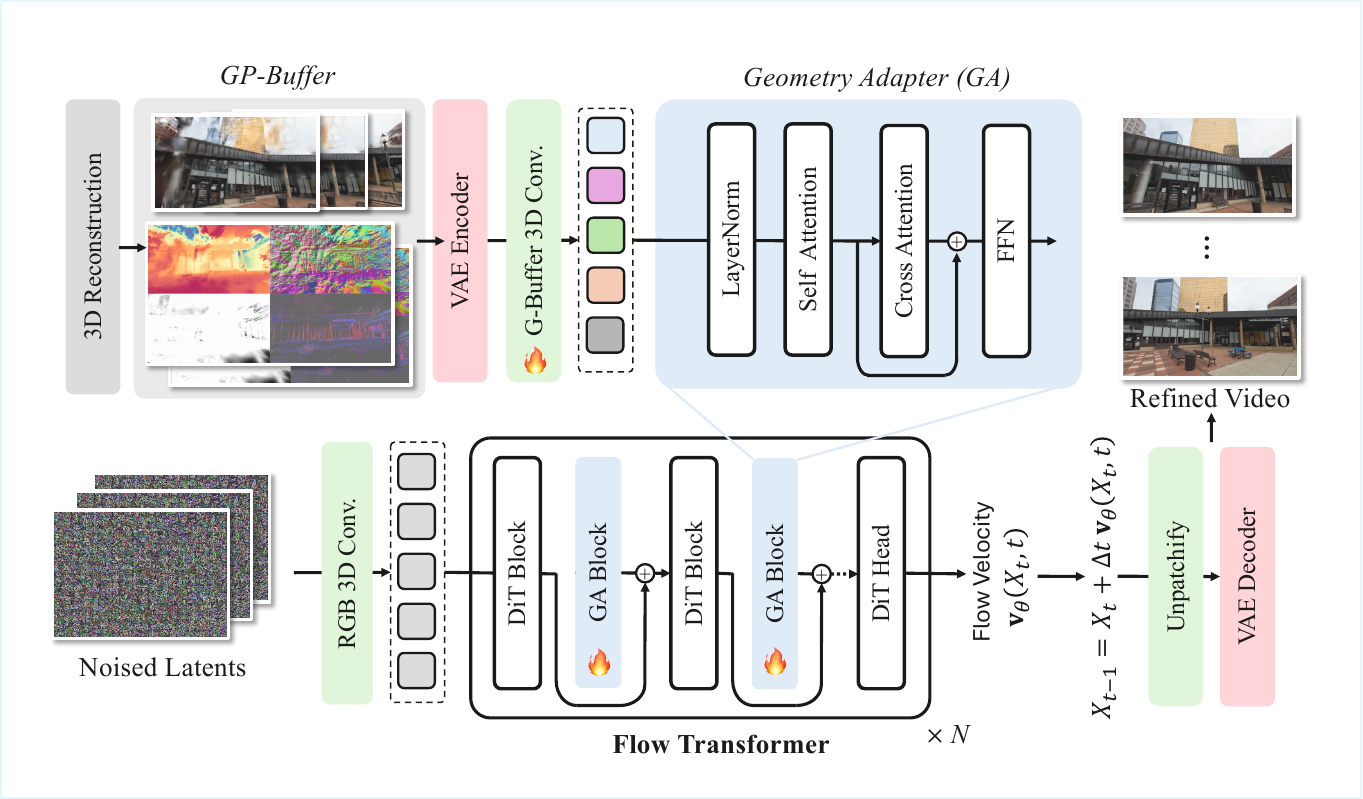}
    \caption{\textbf{\ours Video Generator Architecture.} Our model refines video latents using geometry-aware conditioning derived from 3D Gaussian splatting (3DGS). A Gaussian primitive buffer—comprising color, depth, normals, and uncertainty—is first encoded via a VAE and projected by a 3D convolution into a compact latent $\mathbf{z}_{\mathcal{G}}$. The noised video latents are processed by a flow transformer built upon DiT blocks, interleaved with \textit{Geometry Adapter} (GA) blocks. Each GA block fuses geometry features through self-attention and integrates textual scene descriptions via cross-attention, producing a geometry-aware feature $\mathbf{x}g$ that modulates the video latent $\mathbf{x}$. The model predicts the flow velocity $v\theta(x_t, t)$, which is integrated to recover refined video frames decoded by the VAE.}
    \label{fig:architecture}
\end{figure*}

\subsection{Gaussian Primitives Buffer}
Our key insight is that artifact refinement, encompassing inpainting, outpainting, deblurring, and floater removal, benefits from explicit geometric cues beyond RGB appearance. 
Although directly conditioning on 3D backbones has yielded promise for 3D reconstruction, these methods struggle to scale to large, complex scenes where the number of primitives grows substantially~\cite{chen2024splatformer,liu2024reconx}. To address this, we extract essential 3D Gaussian attributes into a pixel-aligned representation that encodes appearance and geometry:
\begin{equation}
    \mathcal{G} = \{\mathbf{C}, A, D, \mathbf{N}, \mathbf{U}\},
    \label{eq:gbuffer}
\end{equation}
where $\mathbf{C}$, $A$, and $D$ denote the rendered color, opacity, and depth maps; 
$\mathbf{N}$ represents surface normals; 
and $\mathbf{U}$ encodes the inverse projected covariance of each splat, providing a pixel-aligned measure of local geometric uncertainty.
Each of the modalities in $\mathcal{G}$ belongs to $\mathbb{R}^{W \times H \times \{1,3\}\times T }$.
Inspired by the classical \emph{G-buffer} in~\cite{liang2025diffusionrenderer}, we refer to this representation as the \textit{Gaussian Primitives Buffer (GP-Buffer)}.

\paragraph{Alpha, Color, and Depth.}
The color map $\mathbf{C}$ and opacity map $A$ are directly obtained from the standard 3DGS rasterization process. 
Depth is rendered using the \texttt{ED} (expected depth) mode of the renderer~\cite{ye2025gsplat}, which computes the weighted average depth of all contributing splats. 

\paragraph{Normals.}
Screen-space normals are estimated from the camera-space position map $\mathbf{P}_{\text{cam}}$ via finite differences:
\begin{equation}
\mathbf{N}(\mathbf{u}) = \operatorname{normalize}\!\left(\partial_u \mathbf{P}_{\text{cam}} \times \partial_v \mathbf{P}_{\text{cam}}\right),
\end{equation}
where $\partial_u$ and $\partial_v$ denote finite differences along image axes. 
To prevent mixing of foreground and background at silhouettes, normals are evaluated only for pixels with sufficient opacity ($A(\mathbf{u}) > \tau$). 
\vspace{-5pt}
\paragraph{Geometric Uncertainty (Inverse Covariance).}
To provide the model with local geometric uncertainty, we render a map of the inverse projected 2D covariances. 
For each Gaussian, the $2\times2$ symmetric inverse covariance $(\mathbf{\Sigma}_i')^{-1}$ is parameterized by its unique elements 
$\mathbf{u}_i = [a_i, b_i, c_i]$. 
A lightweight rasterization pass accumulates these vectors using $\alpha$-blending, yielding a three-channel ``uncertainty" map 
$\mathbf{U}(\mathbf{u}) \in \mathbb{R}^3$ that encodes per-pixel anisotropy, which we interpret as a measure of structural regularity in splats. 
Low-texture regions are typically represented by a few large Gaussian primitives, resulting in low rendered values, whereas well-captured, high-frequency areas exhibit higher values. Please refer to Supp. for GP-Buffer visualizations.

\subsection{Geometry-Informed Video Generator}
\label{sec:geometry_adapter}
\ours takes as input the GP-Buffer $\mathcal{G}$, optionally conditioned on text descriptions of the scene, and predicts refined color videos (Fig.~\ref{fig:architecture}).
The GP-Buffer provides geometry- and appearance-aware cues extracted from the 3D reconstruction, which are encoded and injected into the generation process through our Geometry Adapter.

\paragraph{Encoding GP-Buffer.}
The video of each modality in $\mathcal{G}$ is first normalized to $[-1, 1]$ and then passed through the encoder separately to obtain per-modality video latents  
$\{\mathbf{z}_{\mathbf{C}}, \mathbf{z}_{A}, \mathbf{z}_{D}, \mathbf{z}_{\mathbf{N}}, \mathbf{z}_{\mathbf{U}}\}$,  
where $\mathbf{z}_{m} \in \mathbb{R}^{\frac{W}{8} \times \frac{H}{8} \times C \times \frac{T}{4}}$ for each modality $m$.  
These latents are concatenated channel-wise to form a unified representation $\mathbf{z}_{\mathcal{G}}:$
\begin{equation}
    \operatorname{concat}(\mathbf{z}_{\mathbf{C}}, \mathbf{z}_{A}, \mathbf{z}_{D}, \mathbf{z}_{\mathbf{N}}, \mathbf{z}_{\mathbf{U}}) 
    \in \mathbb{R}^{\frac{W}{8} \times \frac{H}{8} \times 5C \times \frac{T}{4}}.
\end{equation}
Although the VAE was originally designed for RGB, we find that it can also reconstruct other modalities with less than 1\% relative error (see Supp.). $\mathbf{z}_{\mathcal{G}}$ is then fed to our geometry adapter to guide the generation of the clean video.

\paragraph{Geometry Adapter (GA)} integrates the encoded GP-Buffer representation $\mathbf{z}_{\mathcal{G}}$ into the flow matching process through spatially aligned modulation. 
Within the DiT~\cite{peebles2023scalable} backbone (\cf\cref{fig:architecture}), GA blocks operate as a parallel side-network that extracts hierachical geometric features to condition the main video generation stream. Each GA block takes as input the geometry features derived from the GP-Buffer and injects them into the corresponding DiT layer.
Specifically, the encoded GP-Buffer $\mathbf{z}{\mathcal{G}}$ is projected via 3D convolution to align spatial and channel dimensions:

\begin{align}
\mathbf{z}_{\mathcal{G}}' &= \operatorname{Conv3D}(\mathbf{z}_{\mathcal{G}}), \nonumber \\
\tilde{\mathbf{z}}_{\mathcal{G}} &= 
\operatorname{SelfAttn}(\operatorname{LayerNorm}(\mathbf{z}_{\mathcal{G}}')), \nonumber \\
\mathbf{x}_g &= 
\operatorname{FFN}\!\Big(
\operatorname{CrossAttn}(\tilde{\mathbf{z}}_{\mathcal{G}}, \mathbf{x}_{\text{text}})
\Big), \nonumber\\
\mathbf{x} &\leftarrow \mathbf{x} + \mathbf{x}_g, \nonumber
\label{eq:ga_block}
\end{align}
where $\mathbf{x}_{\text{text}}$ are text-encoder tokens and $\mathbf{x}$ is subsequently fed to the next DiT block.
This allows the model to progressively refine intermediate latents with geometry- and text-conditioned cues.
During training, the model is supervised to predict clean video latents from corrupted ones, where the supervision target corresponds to clean capture latents. 
We use the same flow-matching objective as described in Eq.~\ref{eq:fm}, with the conditioning variable $c$ being our GP-buffer latents $\mathbf{z}_{\mathcal{G}}$.
During inference, the model iteratively predicts velocity fields to produce clean videos.

\subsection{3D Reconstruction Updating}
\label{subsec:refining_3d}
To refine an existing 3D reconstruction, we generate smooth novel-view trajectories that densely sample the camera path between the captured views, following~\cite{fischer2025flowr,Wu2025GenFusion}. We employ an spline-based interpolation scheme that regularizes both camera translation and rotation to ensure physically plausible motion. 
Each camera pose is decomposed into position, look-at, and up vectors, which serve as control points of a B-spline curve, producing a continuous and stable trajectory. 
We rasterize the GP-Buffer (Sec.~\ref{sec:geometry_adapter}) along this trajectory to obtain novel renderings, which are then refined by our geometry-aware video generator to produce artifact-free frames. 
Finally, we merge the generated novel views with the original inputs and optimize the 3D Gaussian splats using the standard photometric loss (Eq.~\ref{eq:loss}), improving both geometric consistency and texture fidelity.

\begin{table*}[!t]
\centering
\small % or \scriptsize
\resizebox{0.95\textwidth}{!}{
\begin{tabularx}{\textwidth}{lc|cccc|cccc|c}
\toprule
 & &\multicolumn{4}{c|}{{\textbf{DL3DV}}} & \multicolumn{4}{c|}{{\textbf{RE10K}}} & \\ 
\midrule
\textbf{Method}& \textbf{Venue} & PSNR$\uparrow$ & SSIM$\uparrow$ & LPIPS$\downarrow$ & FID $\downarrow$& PSNR$\uparrow$ & SSIM$\uparrow$ & LPIPS$\downarrow$ & FID $\downarrow$& Infer. Speed \\
\midrule
Splatfacto~\cite{ye2025gsplat}& {\scriptsize SIGG. 2023} & 17.417 & 0.605 & 0.412 & 6.488 & 19.234 & 0.708 & 0.457 & 12.623 &  (118.3 FPS)\\
GenFusion~\cite{Wu2025GenFusion} & {\scriptsize CVPR 2025} & 18.363 & 0.690 & 0.391 & 9.981 & 20.618 & 0.796 & 0.344 & 14.747 & 1.1 FPS\\
Difix3D+~\cite{wu2025difix3d} & {\scriptsize CVPR 2025} & 20.095 & 0.765 & 0.302 & \rd 4.217 & 24.347 & 0.868 & 0.275 & 13.994 &  \nd12.8 FPS\\
ExploreGS~\cite{Kim_2025_ExploreGS}& {\scriptsize ICCV 2025} &20.689 & 0.760& 0.345 &6.269 & 24.025&0.874 & 0.272& 15.629  & 1.2 FPS\\
\hdashline
\textit{Ours (Single)} & {\scriptsize CVPR 2026}&\rd22.266 & \rd0.827 & \nd0.282 & \nd4.171 & \rd24.754 & \rd0.897 & \rd0.215 & \nd7.496 &  
\rd4.3 FPS
\\
\textit{Ours (Full)} &{\scriptsize CVPR 2026} &\fs22.548 & \nd0.832 & \fs0.278 & \fs3.933 & \fs28.652 & \nd0.944 & \fs0.180 & \fs6.672& \rd4.3 FPS\\
\textit{Ours (Few-step)} & {\scriptsize CVPR 2026}&\nd22.494 & \fs0.842 & \rd0.288 & 7.383 & \nd28.254 & \fs0.952 & \nd0.184 & \rd10.034 &  \fs15.11 FPS\\
\bottomrule
\end{tabularx}
}
\caption{\textbf{Rendering Refinement Performance on DL3DV and RE10K Datasets.} We compare our method against state-of-the-art 3DGS refinement approaches. The joint training variant achieves the best overall fidelity and perceptual quality (highest PSNR/SSIM, lowest LPIPS/FID), while the distilled model attains comparable performance with significantly improved runtime efficiency, reaching 15 FPS.}
\label{tab:recon_perf}
\end{table*}

\subsection{Finetuning Strategy}
Typical video diffusion models require 30–50 iterative denoising steps, resulting in inference times of several minutes per video. To achieve efficient refinement, we adopt a two-stage finetuning strategy that converts a multi-step video generator into a compact four-step refinement model. 
\paragraph{Base Model Distillation.}
We begin with a pretrained text-to-video model and apply distribution matching distillation (DMD)~\cite{yin2024dmd,yin2024onestep} to obtain a 4-step generator. DMD distills a slow, multi-step teacher generator into an efficient few-step student by minimizing the reverse KL divergence between the teacher’s data distribution and the student’s generated distribution across randomly sampled timesteps. The training combines a distribution matching loss~\cite{yin2024onestep} that encourages the student to approximate the teacher’s score distribution, and a standard flow loss in \cref{eq:fm}. For more details on DMD, we refer the readers to~\cite{yin2025causvid}.
\paragraph{Adapter Finetuning.}
After obtaining the distilled generator, we integrate our Geometry Adapter (GA) module into its DiT backbone to get the full model.
All parameters of the distilled model are frozen, and only the GA layers are finetuned (\cf\cref{fig:architecture}).
Note in this stage we \textit{only} use the flow matching loss without DMD, as no further step distillation is required. The timestep schedule of the full model follows the same setup as the distilled model. 

\section{Artifact Simulation}
\label{sec:data}
A key factor in training a generalizable rendering refinement model is a comprehensive data generation pipeline that accurately mimics the diverse artifacts observed in real world. 
We construct paired videos of \emph{ground-truth} and \emph{corrupted} renderings by rendering each scene from two versions of its 3DGS reconstruction: a \textit{high-quality} model from dense inputs and a \textit{corrupted} one with degradation.
\begin{itemize}
    \item \textbf{Sparse-View Simulation.} We randomly retain only $5\%$ of the original video frames to simulate under-sampled captures. 
    This random downsampling, unlike the uniform scheme used in~\cite{wu2025difix3d,Kim_2025_ExploreGS,liu2024dgsenhancer}, introduces temporal irregularity that better reflects real-world conditions.
    \item \textbf{Diverse Initialization.} We apply multiple 3D Gaussian initialization strategies, including structure-from-motion (SfM) point initialization, random 3D point cloud initialization, and dense point maps from MapAnything~\cite{keetha2025mapanything}.
    \item \textbf{Paired Reconstruction with New Trajectories.} A \textit{clean splat} model is trained on all views, while a \textit{corrupted} model uses the sparse subset with fewer optimization steps to simulate underfitting.
Novel camera paths are sampled to synthesize realistic motion artifacts (\cref{subsec:refining_3d}).
    \item \textbf{Feed-forward Degradation.} In addition to optimization-based artifacts, we further render degraded videos using Gaussians predicted by a pretrained feed-forward 3DGS model~\cite{xu2024depthsplat}, introducing new artifacts including geometric inconsistencies, color shifts, and semi-transparent splats. 
    % The original complete captures serve as the ground truth.
\end{itemize}

\section{Experiments}
\paragraph{Implementation Details.}
We adopt Wan-2.1-1.3B~\cite{wan2025} as our base model. It contains 1.3B parameters, and our GA introduces additional 0.6B parameters. Our proposed data curation pipeline produced 75K+ paired video samples from the training splits of DL3DV~\cite{ling2024dl3dv} and RE10K~\cite{re10k}, each with 81 frames.  We hold out their test splits for evaluation. We use Gemini-Flash-2.5~\cite{gemini2.5flash} to generate the text description of each video. We train our model on 8 H200 GPUs for 100K steps with a batch size of 8 and a frame resolution of 480$\times$832. Training uses the AdamW optimizer with a linear learning rate (LR) warm-up over the first 1K steps, followed by a constant LR of 1$\times$10$^{-5}$. We randomly drop out geometry modalities and text input during training.

\paragraph{Experimental Setup.} We evaluate our method on common novel-view synthesis benchmarks: DL3DV~\cite{ling2024dl3dv} and RE10K~\cite{re10k}.
Testing scenes are drawn from the official test splits of each dataset, which remain unseen during training. 
For DL3DV and RE10K, we randomly downsample 5\% of the original video frames as training views and optimize Gaussian splats for 7K iterations following the standard Splatfacto~\cite{ye2025gsplat} setup. The remaining views are held out for evaluation. All methods are tested at their native operating resolution and resized to 480$\times$832 for comparison.

\paragraph{Baselines and Metrics.}
We compare our method against open-source, state-of-the-art approaches, including  DiFiX3D+~\cite{wu2025difix3d},  GenFusion~\cite{Wu2025GenFusion}, and ExploreGS~\cite{Kim_2025_ExploreGS}. We also compare against MVSplat360~\cite{chen2024mvsplat360} on feed-forward view synthesis, a model coupled with MVSplat~\cite{chen2024mvsplat}. Each baseline employs slightly different camera sampling strategies and reconstruction heuristics, making it difficult to disentangle improvements due to model design from those due to optimization choices. 
To ensure a fair comparison, we run all baselines on the same renderings and apply the same reconstruction update strategy described in \cref{subsec:refining_3d}. 
All baselines are evaluated using their official implementations and publicly released weights. 
We assess performance using standard image and perceptual quality metrics: PSNR, SSIM~\cite{ssim}, LPIPS~\cite{lpips}, and FID~\cite{NIPS2017_8a1d6947}. In the following, best results are highlighted as\colorbox{colorFst}{\bf first},\colorbox{colorSnd}{second}, and\colorbox{colorTrd}{third}.

\begin{table}[t]
\centering
\footnotesize
\setlength{\tabcolsep}{4pt}
\begin{tabular}{lcccc}
\toprule
\textbf{Feed-forward Method} & \textbf{PSNR $\uparrow$} & \textbf{SSIM $\uparrow$} & \textbf{LPIPS $\downarrow$} & \textbf{FID $\downarrow$}\\
\midrule
DepthSplat~\cite{xu2024depthsplat} & \nd21.773 & \nd0.842 & \nd0.278  &\rd28.903\\
\quad \textit{w/ }Difix3D+~\cite{wu2025difix3d} & \rd21.415 & \rd0.837 & \rd0.303 & \nd17.968 \\
\quad \textit{w/ }ExploreGS~\cite{Kim_2025_ExploreGS} & 21.195 & 0.816 & 0.317 & 35.475\\
\quad \textit{w/ }\ours(Ours) & \fs 22.802 & \fs 0.848 & \fs0.258 & \fs15.128 \\
\midrule
MVSplat~\cite{chen2024mvsplat} & \rd18.744 & \rd0.706 & \rd0.375 & 44.479\\
% \quad \textit{w/ }MVSplat360~\cite{chen2024mvsplat360} & & & & \\
\quad \textit{w/ }Difix3D+~\cite{wu2025difix3d} & 18.261 & 0.702 & 0.385 & \nd23.954\\
\quad \textit{w/ }ExploreGS~\cite{Kim_2025_ExploreGS} & 18.112 & 0.692 & 0.389 & \rd36.390\\
\quad \textit{w/ }MVSplat360~\cite{chen2024mvsplat360} & \fs19.819&\fs0.747 & \nd0.354& 36.821\\
\quad \textit{w/ }\ours(Ours)  & \nd19.756 & \nd0.744 & \fs0.338 & 
\fs23.823\\
\bottomrule
\end{tabular}
\caption{\textbf{Feed-forward View Synthesis Results on RE10K~\cite{re10k}.} Our method consistently improves feed-forward 3DGS reconstruction on different backbones while baseline methods Difix3D and ExploreGS decrease the consistency with the original scene. MVSplat360 is on par to our method in terms of consistency within its specialized domain, but still lacks in visual quality.}
\label{tab:feedforward_refine}
\end{table}
 
\begin{table}[t]
\centering
\footnotesize
    \centering
    \setlength{\tabcolsep}{1pt}
    \begin{tabular}{lcccccc}
    \toprule
     & \multicolumn{3}{c}{\textbf{DL3DV}} 
     & \multicolumn{3}{c}{\textbf{RE10K}} \\
     \midrule
     \textbf{Method} 
     & PSNR $\uparrow$ & SSIM $\uparrow$ & LPIPS $\downarrow$ 
     & PSNR $\uparrow$ & SSIM $\uparrow$ & LPIPS $\downarrow$ \\
    \midrule
    Splatfacto~\cite{ye2025gsplat} & 17.42 & 0.605 &	0.412 &	19.23 & 0.708 &	0.457 \\
    GenFusion~\cite{Wu2025GenFusion} & 18.08 &	\rd0.615 &	 0.421&	19.52 &	0.719 & 0.329 \\
    Difix3D+~\cite{wu2025difix3d} & \nd18.87 & \nd0.626 &	\fs0.265 & \rd 21.98	 &	\nd0.767 &	\nd0.187 \\
    ExploreGS~\cite{Kim_2025_ExploreGS} &\rd18.71 & 0.611& \rd0.377&\nd22.06 & \rd0.742 & \rd0.257\\
    \ours (Ours) & \fs19.36 &	\fs0.628 &	\nd0.279 &	\fs24.17 & \fs0.817 & \fs0.175 \\
    
    \bottomrule
    \end{tabular}
    \caption{\textbf{Performance on Improving 3D Reconstruction}. \ours demonstrates superior performance on most metrics, showcasing the multi-view consistency in our enhanced frames. }
    \label{tab:splat_refine}
\end{table}

\begin{figure*}[t]
    \centering
 \begin{tabular}{cccccc}
   
    \multicolumn{6}{c}{%
        \includegraphics[width=\linewidth]{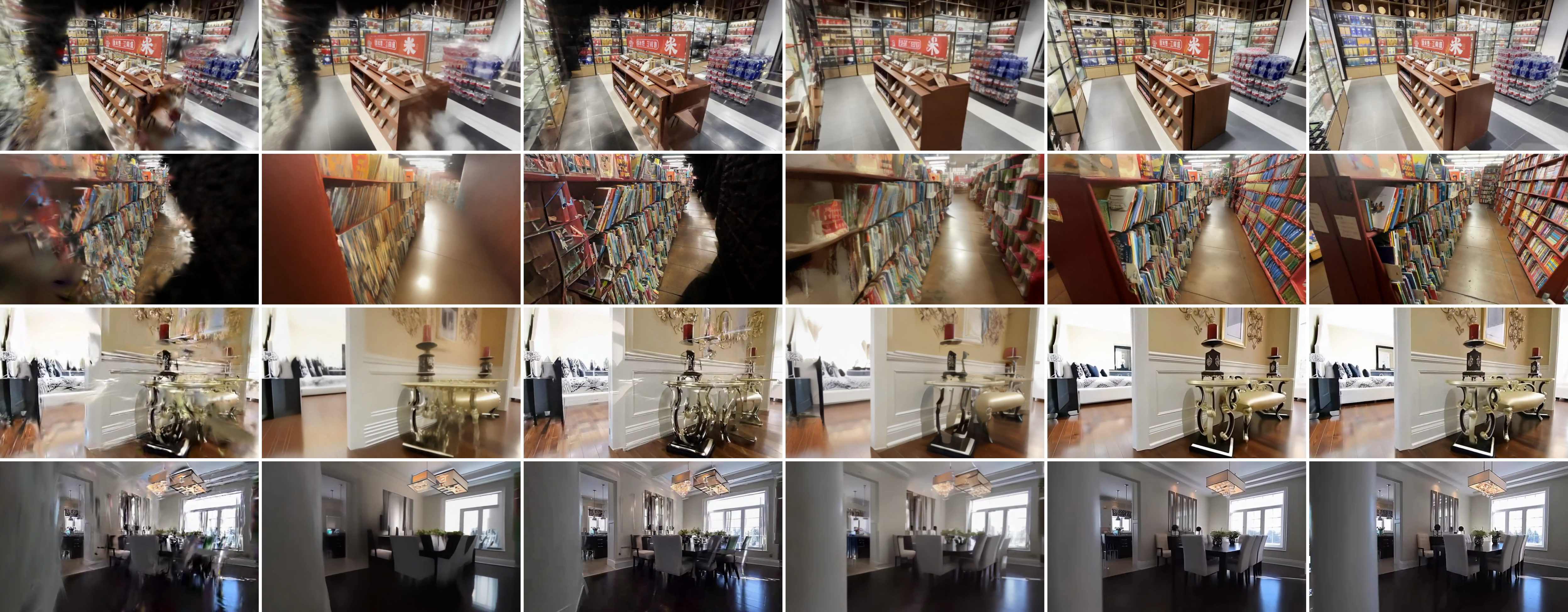}%
    } \\
     \hspace{0.4cm} \small Splatfacto~\cite{ye2025gsplat} & 
    \hspace{0.4cm} \small GenFusion~\cite{Wu2025GenFusion} & 
    \hspace{.5cm}\small Difix3D+~\cite{wu2025difix3d} & 
    \hspace{.6cm}\small ExploreGS~\cite{Kim_2025_ExploreGS} &
    \hspace{0.2cm}\small \ours (Ours) & 
    \hspace{-0.1cm}\small Ground Truth \\
    \midrule
    \multicolumn{6}{c}{%
        \includegraphics[width=\linewidth]{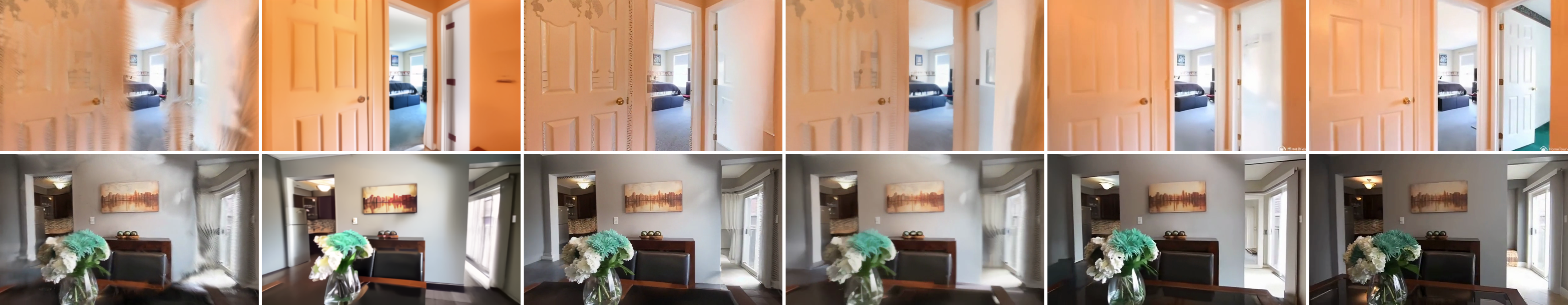}%
    } \\
    \hspace{0.4cm} \small DepthSplat~\cite{xu2024depthsplat} & 
    \hspace{0.4cm} \small MVSplat360~\cite{chen2024mvsplat360} & 
    \hspace{.5cm}\small Difix3D+~\cite{wu2025difix3d} & 
    \hspace{.6cm}\small ExploreGS~\cite{Kim_2025_ExploreGS} &
    \hspace{0.2cm}\small \ours (Ours) & 
    \hspace{-0.1cm}\small Ground Truth \\

\end{tabular}  
   \caption{\textbf{Qualitative Comparison on Novel-View Refinement.}
We compare \ours with baseline methods on diverse scenes from DL3DV~\cite{ling2024dl3dv} and RE10K~\cite{re10k}. 
\ours effectively removes rendering artifacts such as blur, floaters, ghosting, and texture distortions, producing sharper geometry, cleaner reconstruction than Splatfacto~\cite{ye2025gsplat}, GenFusion~\cite{Wu2025GenFusion}, DiFiX3D+~\cite{wu2025difix3d}, and ExploreGS~\cite{Kim_2025_ExploreGS}, and achieving visual fidelity closest to the ground truth. 
From top to bottom, our model demonstrates strong generalization across various refinement scenarios: (\textit{1}) inpainting missing regions, (\textit{2}) outpainting beyond the original view, (\textit{3–4}) correcting blur, ghosting, and geometric errors, and (\textit{5–6}) handling artifacts from feed-forward 3DGS models, including transparent Gaussian primitives and spiky geometry. 
% While baselines handle certain artifact types, they fail to generalize as robustly as \ours.
}

    \label{fig:image_refine}
\end{figure*}

\subsection{Results}
\paragraph{Improving Novel-View Synthesis.}
We first evaluate performance on pure rendering refinement, \ie, novel-view synthesis, as reported in Table~\ref{tab:recon_perf}. We compare three variants of our model: \textit{Ours (Single)}, trained solely on DL3DV~\cite{ling2024dl3dv} with optimization-based data; \textit{Ours (Joint)}, jointly trained on all datasets with the complete artifact data; and \textit{Ours (Distilled)}, our distilled 4-step model. The model trained exclusively on DL3DV outperforms all baselines trained on the same dataset by a substantial margin in terms of image quality.

Difix3D+~\cite{wu2025difix3d}, based on an image diffusion model, processes frames independently and thus lacks multi-view consistency and performs poorly on removing floaters. GenFusion~\cite{Wu2025GenFusion} and ExploreGS~\cite{Kim_2025_ExploreGS}, though leveraging a video diffusion framework, achieve lower performance due to insufficient conditioning signals compared to our GP-buffer and sub-optimal conditioning strategy, which impairs temporal coherence. More importantly, we find our joint trained model has improved performance compared to training on a single dataset, indicating that exposure to diverse reconstruction artifacts and scene types enhances the model’s performance on the same dataset, demonstrating the importance of mix training across different datasets and methods.
\cref{fig:image_refine} further confirms our method's superior performance: the four different examples examine the model's ability in inpainting, outpainting, sharpening the blurry, and correcting wrong geometry in the original rendering.

\paragraph{Improving Feed-Forward View Synthesis.}
We evaluate our model’s ability to refine renderings produced by feed-forward 3DGS reconstruction methods, DepthSplat~\cite{xu2024depthsplat} and MVSplat~\cite{chen2024mvsplat}, as reported in \cref{tab:feedforward_refine}. Prior refinement approaches generally struggle on feed-forward reconstructions—MVSplat360~\cite{chen2024mvsplat360} being the sole exception, as it is explicitly tailored for MVSplat. In contrast, \ours generalizes effectively across different feed-forward backbones.
Although our method is evaluated on MVSplat outputs without ever being trained on MVSplat predictions, it performs on par with MVSplat360, which is fully trained on that setting. Qualitative results in \cref{fig:image_refine} further confirm that our method can reliably correct spiky primitives and semi-transparent Gaussian artifacts.

\vspace{-10pt}

\paragraph{Inference Speed.}
We report inference speed in \cref{tab:recon_perf}, measuring the average frame rate based on end-to-end inference time on a single H200 GPU. Our distilled model maintains comparable PSNR, SSIM, and LPIPS to the non-distilled version, while surpassing all baselines with a real-time speed of 16~FPS. A slightly higher FID score is observed, which we attribute to the reduced number of denoising steps and minor loss of high-frequency details.

\begin{figure}[t]
    % \centering
    \includegraphics[width=\linewidth]{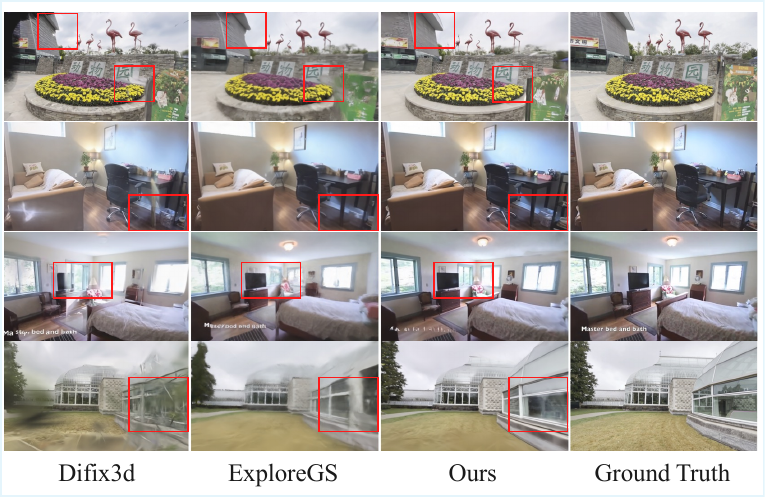}%
   \caption{\textbf{Qualitative Comparison on 3D Reconstruction.} We show the novel-view renderings from the improved 3D reconstruction refined using enhanced views from different methods.
   \ours consistently outperforms baseline methods. }
    \label{fig:recon_comparison}
\end{figure}

\paragraph{Improving 3D Reconstruction.}
We evaluate the effectiveness of \ours in enhancing the underlying 3DGS representation rather than only refining individual frames. 
This task jointly measures rendering fidelity and 3D consistency, with metrics reported per scene to assess holistic reconstruction quality. 
As shown in \cref{tab:splat_refine}, \ours achieves superior PSNR and SSIM across datasets, with occasionally slightly higher LPIPS than Difix3D due to its perceptual loss during training. 
Overall, \ours produces sharper, cleaner, and more consistent reconstructions, significantly improving both rendering quality and geometric coherence (see \cref{fig:recon_comparison}).

\subsection{Ablation Studies}
We validate our key design choices from three perspectives: input modalities, training data, and model architecture.
\vspace{-5mm}

\begin{table}[t]
\centering
\scriptsize
\setlength{\tabcolsep}{3pt}
\begin{tabular}{ccccc|ccccc}
\toprule
\multicolumn{5}{c}{\textbf{Input Modality}} & \multicolumn{4}{c}{\textbf{Performance}} \\
RGB & Depth & Normal & Alpha & Cov. & PSNR$\uparrow$ & SSIM$\uparrow$ & LPIPS$\downarrow$ & FID$\downarrow$ \\
\midrule
\checkmark &  &  &  &  & 19.148 & 0.719 & 0.385 & 15.451 \\
\checkmark & \checkmark &  &  &  & 19.289 & 0.726 & 0.361 & 10.542 \\
\checkmark & \checkmark & \checkmark &  &  & 19.744 & 0.738 & 0.355 & 10.291 \\
\checkmark & \checkmark & \checkmark & \checkmark &  & 19.958 & 0.748 & 0.344 & 8.612 \\
\checkmark & \checkmark & \checkmark & \checkmark & \checkmark & \fs20.753 & \fs0.773 & \fs0.329 & \fs6.724 \\
\bottomrule
\end{tabular}
\caption{\textbf{Ablation on Input Modalities on DL3DV Dataset.} Inclusion of more geometric cues (depth, normal, alpha, covariance) improves reconstruction fidelity and perceptual quality. 
}
\label{tab:ablation_modalities}
\end{table}

\begin{figure}[t]
    \centering
    \includegraphics[width=.7\linewidth]{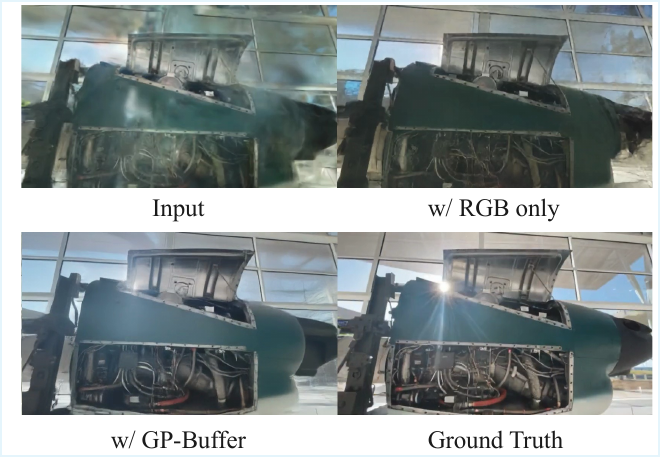}
    \caption{\textbf{Qualitative Ablation on GP-Buffer.} Additional geometric attributes help the model “see through” floaters and blur, producing cleaner and sharper results than RGB-only.}
    \label{fig:gbuffer}
\end{figure}

\paragraph{Input Modality.} We analyze the contribution of each geometric channel in the GP-Buffer by training five separate models, each with a different combination of input modalities. For efficiency, all models are trained on one-third of the full dataset. As per \cref{tab:ablation_modalities}, adding geometric modalities (depth, normal, alpha, covariance) leads to consistent improvements in rendering accuracy and perceptual quality. \cref{fig:gbuffer} further confirms that GP-buffer can better disambiguate noisy regions and produce sharper reconstructions.

\paragraph{Artifact Simulation (Data).}
To isolate data generation from architecture, we fix the backbone and training setup, and compare prior artifact synthesis~\cite{Kim_2025_ExploreGS,Wu2025GenFusion,wu2025difix3d} only using uniform view downsampling and splat underfitting, against our proposed pipeline (Sec.~\ref{sec:data}), which combines optimization- and feed-forward degradations while injecting pose and coverage diversity. 
Our synthesis yields higher PSNR/SSIM and lower LPIPS/FID (Tab.~\ref{tab:ablation_artifact}, rows 1\&3). Training with a \emph{mixed} artifact set improves both feed-forward \emph{and} optimization-based evaluations (Tab.~\ref{tab:splat_refine}), suggesting that exposure to cross-paradigm artifacts produces a more balanced refiner rather than overfitting to one pipeline.

\begin{table}[t]
\centering
\scriptsize
\setlength{\tabcolsep}{5pt}
\begin{tabular}{lc|cccc}
\toprule
\multicolumn{2}{c|}{\textbf{Training Config.}} & \multicolumn{4}{c}{\textbf{Performance}} \\
Artifact Sim. & Architecture & PSNR$\uparrow$ & SSIM$\uparrow$ & LPIPS$\downarrow$ & FID$\downarrow$ \\
\midrule
Baseline & \textit{w/.} GA & \nd21.087 & \rd0.789 & \nd0.303 & \rd5.717 \\
Ours & \textit{w/o.} GA & \rd20.904 & \nd0.794 & \rd0.308 & \nd5.296 \\
Ours & \textit{w/.} GA & \fs22.548 & \fs0.832 & \fs0.278 & \fs3.933 \\
\bottomrule
\end{tabular}
\caption{\textbf{Ablation on Artifact Simulation and Architecture.} Evaluating the impact of artifact simulation and geometry adapter block on reconstruction fidelity and perceptual quality.}
\label{tab:ablation_artifact}
\end{table}

\paragraph{Geometry Adapter.} 
Existing methods such as GenFusion~\cite{Wu2025GenFusion} and ExploreGS~\cite{Kim_2025_ExploreGS} condition generation by directly adding conditioning latents to the noised latents. 
We find this strategy suboptimal: as shown in \cref{tab:ablation_artifact} (rows 2\&3), replacing our GA with a simple convolutional fusion layer (\textit{w/o GA}) yields slower convergence and weaker geometric alignment between conditioning and target structures. 
In contrast, incorporating our GA consistently improves all metrics, raising PSNR from 20.90 to 22.55 and reducing LPIPS and FID by 0.03 and 1.36, respectively. 
\section{Conclusion}
We present \ours, a geometry-informed video generative model that enhances 3DGS reconstructions by leveraging depth, normals, and uncertainty cues.
It effectively removes floaters, flickering, and blur while preserving fine details.
Trained with a diverse artifact synthesis pipeline, \ours achieves state-of-the-art performance, and its distilled few-step variant enables real-time inference.
We discuss our limitations and future work in Supp.

\paragraph{Acknowledgements.}
We thank Lambert Wixson and Yin Wang for the fruitful discussion.

% \clearpage

{
    \small
    \bibliographystyle{ieeenat_fullname}
    \bibliography{main}
}

\clearpage

\setcounter{page}{1}
\setcounter{section}{0}
\counterwithin{figure}{section} % add section number to table
\counterwithin{table}{section}
\renewcommand{\thesection}{\Alph{section}} % numbering to letters
\maketitlesupplementary
\begin{abstract}
    We provide additional details in the following:
\begin{enumerate}
    \item A website including additional video results (\cref{supp:website}).
    \item Implementation details of \ours, including more training details and model architecture (\cref{supp:implementation}).
    \item Additional details on training data generation (\cref{supp:training_data}).
    \item Finetuning strategies (GA vs DiT) (\cref{supp:finetuning}).
    \item Multi-modal encoding using RGB Video VAE (\cref{supp:vae_encoding}).
    \item Analysis of different 3DGS artifacts
    (\cref{supp:3dgs_artifact}).
    \item Comparison of refinement and direction Generation
    (\cref{supp:refinement_vs_generation}).
    \item Additional qualitative results (\cref{supp:add_results}).
    \item Limitations and future work (\cref{supp:limitations}).
\end{enumerate}
\end{abstract}

% \section{Additional Results}
\section{Website and Video}
\label{supp:website}
We provide additional video results in the \href{http://research.zhuliyuan.net/projects/GaussFusion/}{project page} to comprehensively demonstrate the performance of GaussFusion. The website includes (1) multi-modal GP-buffer visualizations displaying the geometry-aware representations (alpha, color, depth, normal, and inverse covariance maps) that inform our video generation process; (2) novel-view refinement results with side-by-side comparisons against baseline methods; and (3) refinement results on feed-forward reconstructions from DepthSplat~\cite{xu2024depthsplat} with side-by-side comparisons against baselines.

\section{Additional Implementation Details}
\label{supp:implementation}
\paragraph{Architecture Details.}
Our flow transformer follows the Wan DiT~\cite{wan2025} backbone with 30 DiT blocks (hidden dimension 1536) and a final DiT head, as illustrated in Fig.~\ref{fig:architecture}. On top of this backbone we interleave lightweight Geometry Adapter (GA) blocks: after every two DiT blocks we insert one GA block, resulting in 15 GA blocks in total~\cite{vace_wan}. Each GA block operates at the same dimensionality as the DiT backbone and follows the structure shown in Fig.~\ref{fig:architecture}, consisting of LayerNorm, self-attention, cross-attention, and an MLP. The 3DGS GP-buffer (alphas, color, depth, normals, and uncertainty) is encoded by a VAE into compact geometry tokens $\mathbf{z}_{\mathrm{g}}$ which are passed to a 3D convolutional layer, then linearly projected to the DiT hidden size, and used as key/value in the GA attention layer. The GA produces a geometry-aware feature $\mathbf{x}_{\mathrm{g}}$, which is projected back and injected into the DiT stream via a gated residual connection, allowing geometry-conditioned features to modulate the video latents while preserving the original Wan DiT parameters.

\paragraph{Training.} We implement our code using PyTorch 2.7.1 with CUDA 12.6 on 8 NVIDIA H200 GPUs. We employ BF16 precision, FlashAttention-2~\cite{dao2023flashattention}, and ZeRO-2 sharding~\cite{aminabadi2022deepspeed} to improve training and memory efficiency. We set batchsize per GPU to 1 and apply gradient accumulation every 4 steps to have an effective batchsize of 32. The training videos have a spatial resolution of 480$\times$832 and 81 frames. We use Wan2.1~\cite{wan2025} VAE to precompute the video latents of all the training pairs for efficiency. The video latents have a spatial resolution of 60$\times$104 and 21 frames in latent space.

\begin{figure}[h]
    \centering
    \includegraphics[width=\linewidth]{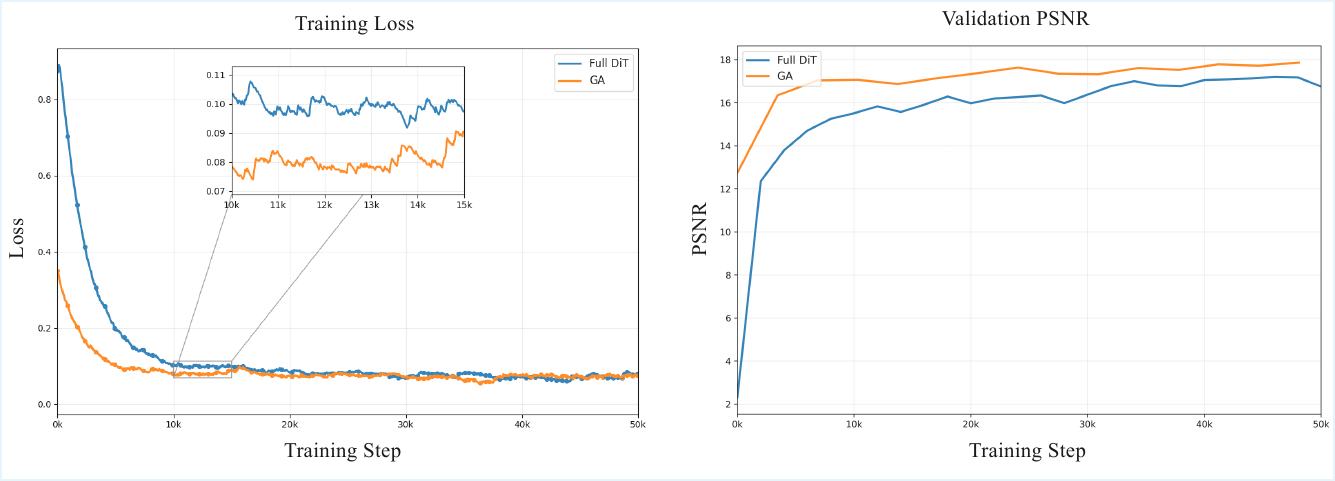}
    \caption{\textbf{Training Curve Comparison Between DiT and GA.} We visualize the training loss (left) and validation PSNR (right) over 50k iterations. The proposed GA method (orange) demonstrates faster convergence and consistently lower loss compared to the Full DiT baseline (blue). Correspondingly, GA achieves higher image restoration fidelity (PSNR) on the validation set, maintaining a clear performance margin throughout same steps of training.}
    \label{fig:training_curve}
\end{figure}

\section{Training Data Generation.}
\label{supp:training_data}
We provide more details of artifact simulation in \cref{sec:data}. We derive the training data on full DL3DV~\cite{ling2024dl3dv} and first 100 training splits (10K scenes) of RE10K~\cite{re10k}. DL3DV provides COLMAP~\cite{schoenberger2016colmap} structure-from-motion (SfM) data so we directly initialize 3DGS from SfM point clouds following the standard protocol. We also adopt random 3D point cloud initialization on 50\% of the scenes in DL3DV. Since RE10K~\cite{re10k} dataset only provides camera poses and does not provide COLMAP reconstruction data, we attempted to run COLMAP ourselves but the default COLMAP reconstruction pipeline failed on most of the RE10K scenes. Therefore we initialize the 3DGS (50\%) from random 3D point initialization and dense point map (50\%) reconstructed from MapAnything~\cite{keetha2025mapanything}. This makes RE10K a harder case because the geometry initialization of the 3DGS includes geometric errors. For optimization-based 3DGS reconstruction, we randomly select training steps from 3000, 7000, and 30000 to imitate undertrained splats.
For feed-forward reconstructions, we run MVSplat~\cite{chen2024mvsplat} and DepthSplat~\cite{xu2024depthsplat} on RE10K dataset to simulate diverse reconstruction artifacts from feed-forward models. For each scene on RE10K, we only feed three evenly distributed views (first, middle, and last) as input to the model to get the sparse-view reconstruction with artifacts.

\section{Geometry Adapter vs DiT Fine-tuning}
\label{supp:finetuning}
To support our claim in the main paper that GA-based finetuning can converge faster and provide better performance than baseline methods under the same number of training steps (see \cref{tab:ablation_artifact}), we provide the training loss and validation curve of DiT-based and GA-based finetuning in \cref{fig:training_curve}. Our GA-based finetuning converges faster in the first 10K training steps and also shows better validation performance than the full DiT-based finetuning. We notice that full DiT-based finetuning can reach similar performance \wrt GA if we train longer for 2$\times$ iterations (3 days). This demonstrates our GA design is more efficient than baselines.

\section{Multi-Modal Encoding}
\label{supp:vae_encoding}
Since the original Wan~\cite{wan2025} VAE was designed to compress RGB videos and remains frozen in our training setup, we study how effective it is to encode non-RGB modalities, \ie, depth, alphas, inverse covariance, and normal maps. We measure the encode-decode reconstruction error for each modality in the GP-Buffer using mean absolute error (MAE), mean squared error (MSE), relative error (RelErr), and squared relative error (SqRel). We run this evaluation on 400 GP-Buffer videos on DL3DV renders and report the metrics in \cref{tab:vae}. We find that depth and alpha have reconstruction errors comparable to RGB, while inverse covariance maps and normal maps exhibit slightly larger errors, likely due to their different color value distributions. This can be alleviated by finetuning the VAE on new modality data following~\cite{jiang2025geo4d,szymanowicz2025bolt3d}.

\begin{table}[h]
\centering
\footnotesize
\setlength{\tabcolsep}{4pt}
\begin{tabular}{lccccc}
\toprule
\textbf{Modality} & \# Channel & \textbf{MAE}$\downarrow$ & \textbf{MSE}$\downarrow$ & \textbf{RelErr}$\downarrow$ & \textbf{SqRel}$\downarrow$ \\
\midrule
RGB  &  3 & 0.0323 & 0.00362 & 0.0199 & 0.0217 \\
Depth&  1 & 0.0253 & 0.00252 & 0.0044 & 0.0182 \\
Alpha&  1 & 0.0138 & 0.00150 & 0.0075 & 0.0053 \\
Covar.& 3 & 0.0758 & 0.01048 & 0.0147 & 0.0847 \\
Normal& 3 & 0.1890  & 0.06684 & 0.1333 & 0.3507 \\
\bottomrule
\end{tabular}
\caption{\textbf{Per-Modality VAE Reconstruction Error.} Comparison of mean absolute error (MAE), mean squared error (MSE), relative error (RelErr), and squared relative error (SqRel).}
\label{tab:vae}
\end{table}

\section{Analysis of 3DGS Artifacts.} 
\label{supp:3dgs_artifact}
Understanding degradation patterns is key to generalizable refinement. While quantitative attribution is challenging due to the stochastic nature of 3DGS optimization (e.g., densification heuristics and camera initialization), we provide our empirical findings and insights in \cref{tab:artifact_comparison}. For \textbf{optimization-based} methods, we observe that large pose drifts generate high-frequency ``floaters,'' while sparse coverage leads to ``needle-like'' over-elongated Gaussians. 
In contrast, \textbf{feed-forward} models~\cite{xu2024depthsplat,chen2024mvsplat} exhibit distinct failure modes driven by regression uncertainty: they suffer from low-frequency blur (due to mean-seeking behavior) and global geometric warping (due to inconsistent depth estimation). Crucially, we do not simulate these artifacts via synthetic filters; instead, we induce them as direct outputs of the 3DGS pipeline by modulating input conditions. This allows us to control the severity of intrinsic failures, ensuring \ours generalizes to practical real-world scenarios.

\begin{table}[h]
\centering
% Resize the table to fit the width of the column
\resizebox{\columnwidth}{!}{%
\begin{tabular}{lll}
\toprule
Feature & Optimization-Based 3DGS & Feed-Forward 3DGS \\ 
\midrule
Primary visual flaw & High-frequency noise (Floaters) & Low-frequency blur (Oversmoothing) \\
Geometry artifacts & "Needles" (over-stretched covariance) & Global depth warping / Scale ambiguity \\
Unseen areas & Empty / Holes & Plausible but blurry (learned priors) \\
Source of error & Overfitting (high variance) & Uncertainty / Regression to mean \\
\bottomrule
\end{tabular}%
}
% \vspace{-pt}
\caption{\textbf{Empirical Analysis of Artifacts in 3DGS.}}
\label{tab:artifact_comparison}
\end{table}

\section{Refinement vs. Direct Generation.} 
\label{supp:refinement_vs_generation}
We evaluate the direct generation baseline, Matrix3D~\cite{lu2025matrix3d}, on DL3DV and find \ours significantly outperforms it (PSNR 19.36 vs. 11.30; LPIPS  0.279 vs. 0.639). Direct generation methods struggle with consistency due to limited context windows and the difficulty of implicitly learning precise 3D projections. Our refinement paradigm uses the 3DGS as an explicit 3D memory. By decoupling the camera model via rasterization, we enforce mathematically correct projections, preventing the drift in pure generation. 

\section{Additional Qualitative Results}
\label{supp:add_results}
In this section, we provide example visuals on the modalities comprising the GP-Buffer input and qualitative comparisons against baseline methods. 

\begin{figure*}[h]
    \centering
    \includegraphics[width=0.8\linewidth]{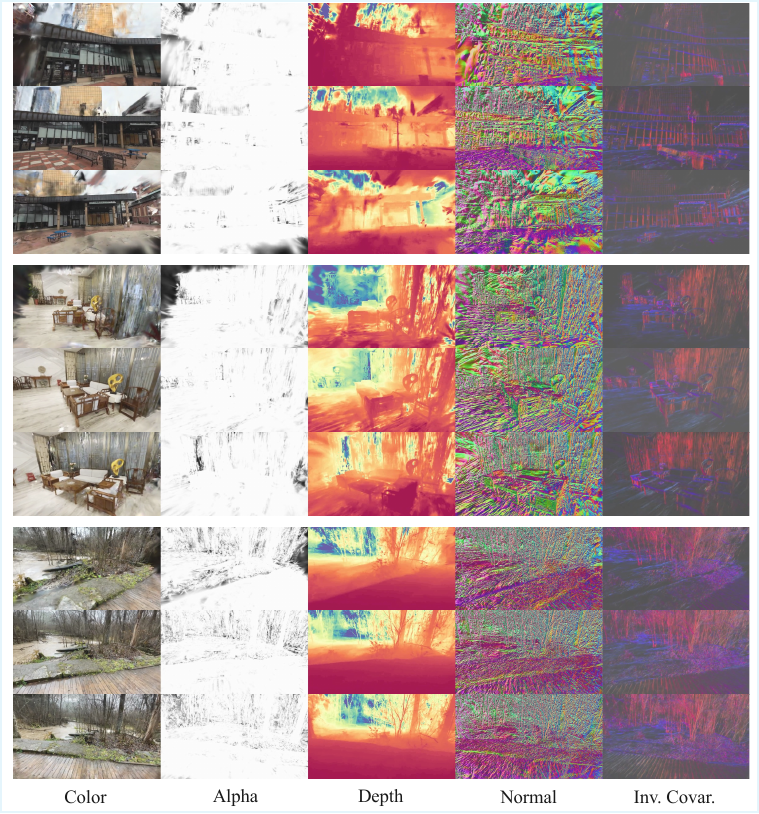}%
\caption{\textbf{GP-Buffer Sample Visualizations on DL3DV~\cite{ling2024dl3dv}.} The five comprising modalities (color, alpha, depth, normal, and inverse covariance) are the input to our model.}
    \label{fig:gp_buffer_vis}
\end{figure*}

\paragraph{GP-Buffer Visualization.}
The GP-Buffer encodes a set of geometry-related modalities, therefore inheriting the characteristic artifacts of 3D Gaussian splatting, that serve as the \emph{input} to our method. As shown in Figure~\ref{fig:gp_buffer_vis}, it contains five modalities: \textit{color}, \textit{alpha}, \textit{depth}, \textit{normal}, and \textit{inverse covariance}. These inputs not only describe the reconstructed scene but also make visible the distortions, noise, and inconsistencies that arise from imperfect geometry or view-dependent effects in the underlying 3DGS representation.
As shown in \cref{fig:gp_buffer_vis}, the color channel reflects photometric artifacts such as blurred textures or multi-view inconsistencies. The alpha channel highlights irregular opacity and over-accumulation in regions where splats overlap or geometry is uncertain. Depth maps often expose depth bleeding and surface discontinuities, while the surface normals reveal noisy or unstable local orientations, especially near edges or thin structures. Finally, the inverse covariance map visualizes the anisotropy and uncertainty of individual Gaussians; high-variance or poorly constrained areas appear as strong patterns in this channel. Together, these complementary modalities make explicit how reconstruction artifacts manifest across the 3D representation, supporting our claim that the GP-Buffer encodes both scene geometry and its associated imperfections.

\begin{figure*}[t]
    \centering
 \begin{tabular}{cccccc}
   
    \multicolumn{6}{c}{%
        \includegraphics[width=\linewidth]{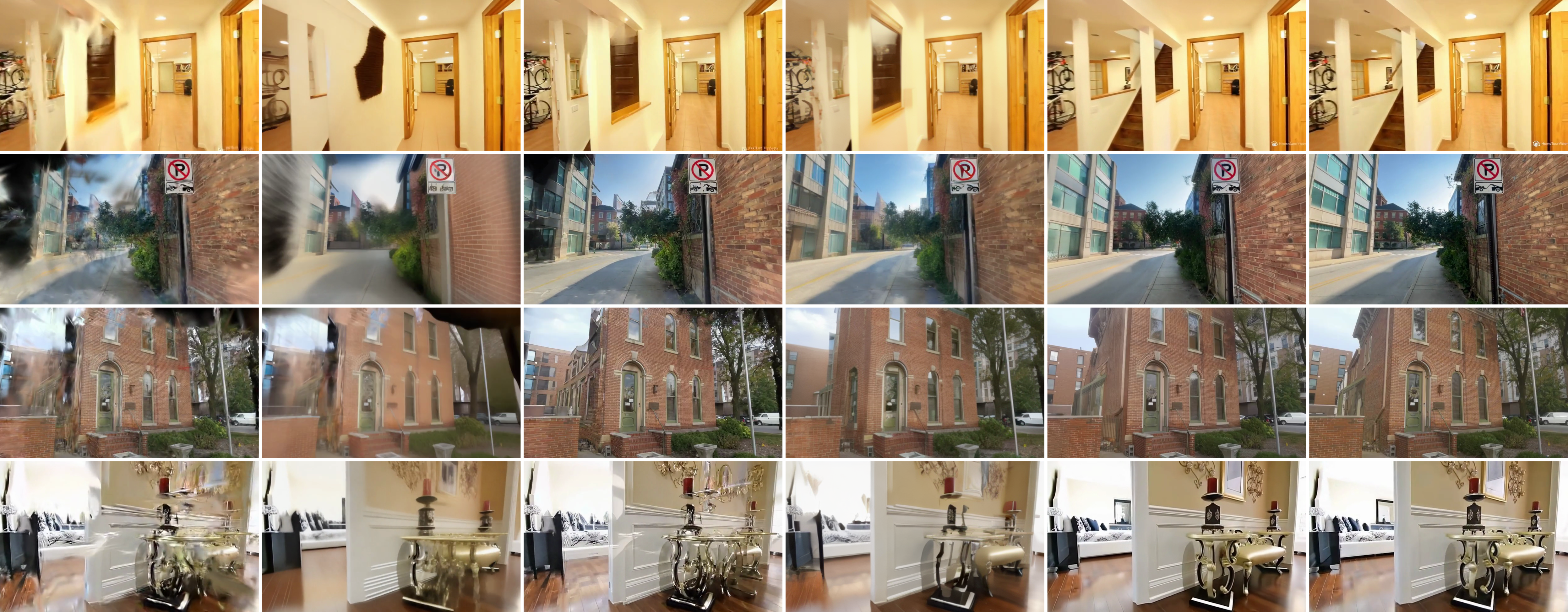}%
    } \\
     \hspace{0.4cm} \small Splatfacto~\cite{ye2025gsplat} & 
    \hspace{0.4cm} \small GenFusion~\cite{Wu2025GenFusion} & 
    \hspace{.5cm}\small Difix3D+~\cite{wu2025difix3d} & 
    \hspace{.6cm}\small ExploreGS~\cite{Kim_2025_ExploreGS} &
    \hspace{0.2cm}\small \ours (Ours) & 
    \hspace{-0.1cm}\small Ground Truth \\
    \midrule
    \multicolumn{6}{c}{%
        \includegraphics[width=\linewidth]{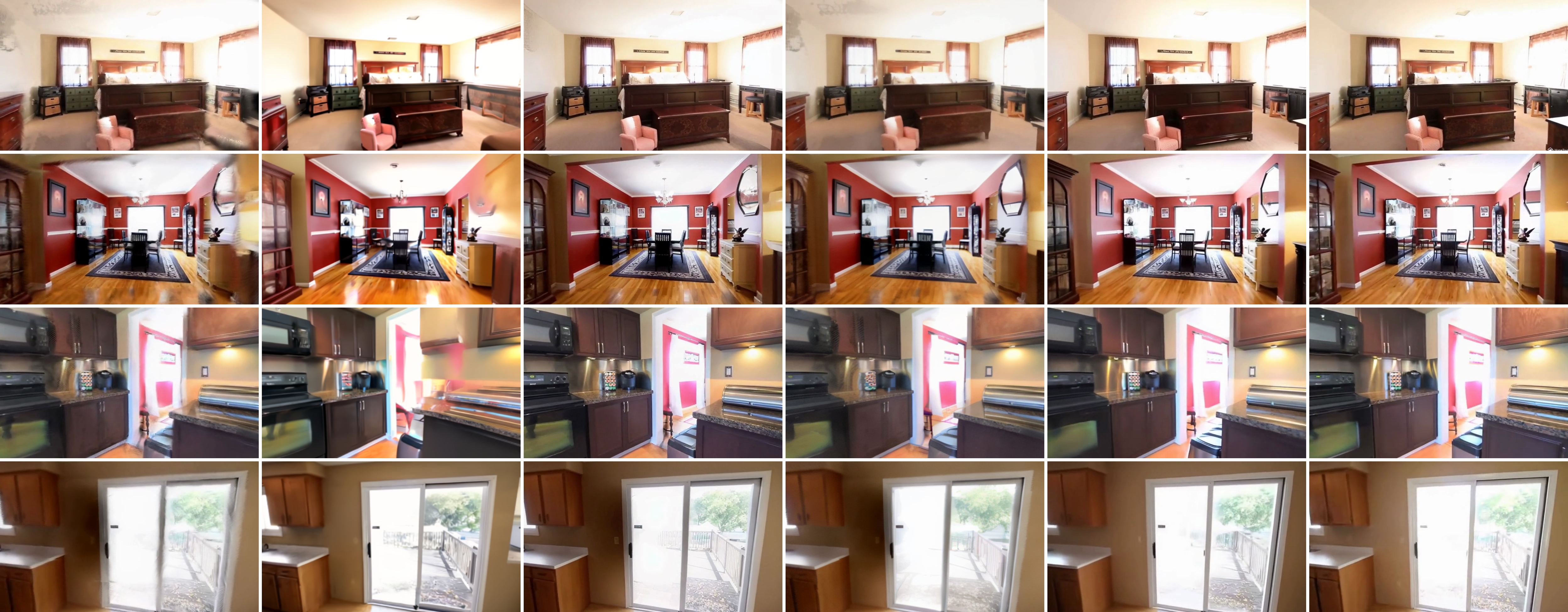}%
    } \\
    \hspace{0.4cm} \small DepthSplat~\cite{xu2024depthsplat} & 
    \hspace{0.4cm} \small MVSplat360~\cite{chen2024mvsplat360} & 
    \hspace{.5cm}\small Difix3D+~\cite{wu2025difix3d} & 
    \hspace{.6cm}\small ExploreGS~\cite{Kim_2025_ExploreGS} &
    \hspace{0.2cm}\small \ours (Ours) & 
    \hspace{-0.1cm}\small Ground Truth \\

\end{tabular}  
   \caption{\textbf{Additional Qualitative Comparison on Novel-view Refinement on DL3DV~\cite{ling2024dl3dv} and RE10K~\cite{re10k}.} Top four rows showcase 3DGS optimization methods and bottom four rows feed-forward 3DGS methods as starting point of the refinement.
}

    \label{fig:supp_image_refine}
\end{figure*}

\begin{figure*}[t]
    \centering

    \begin{tabular}{ccccc}
   
    \multicolumn{5}{c}{%
        \includegraphics[width=\linewidth]{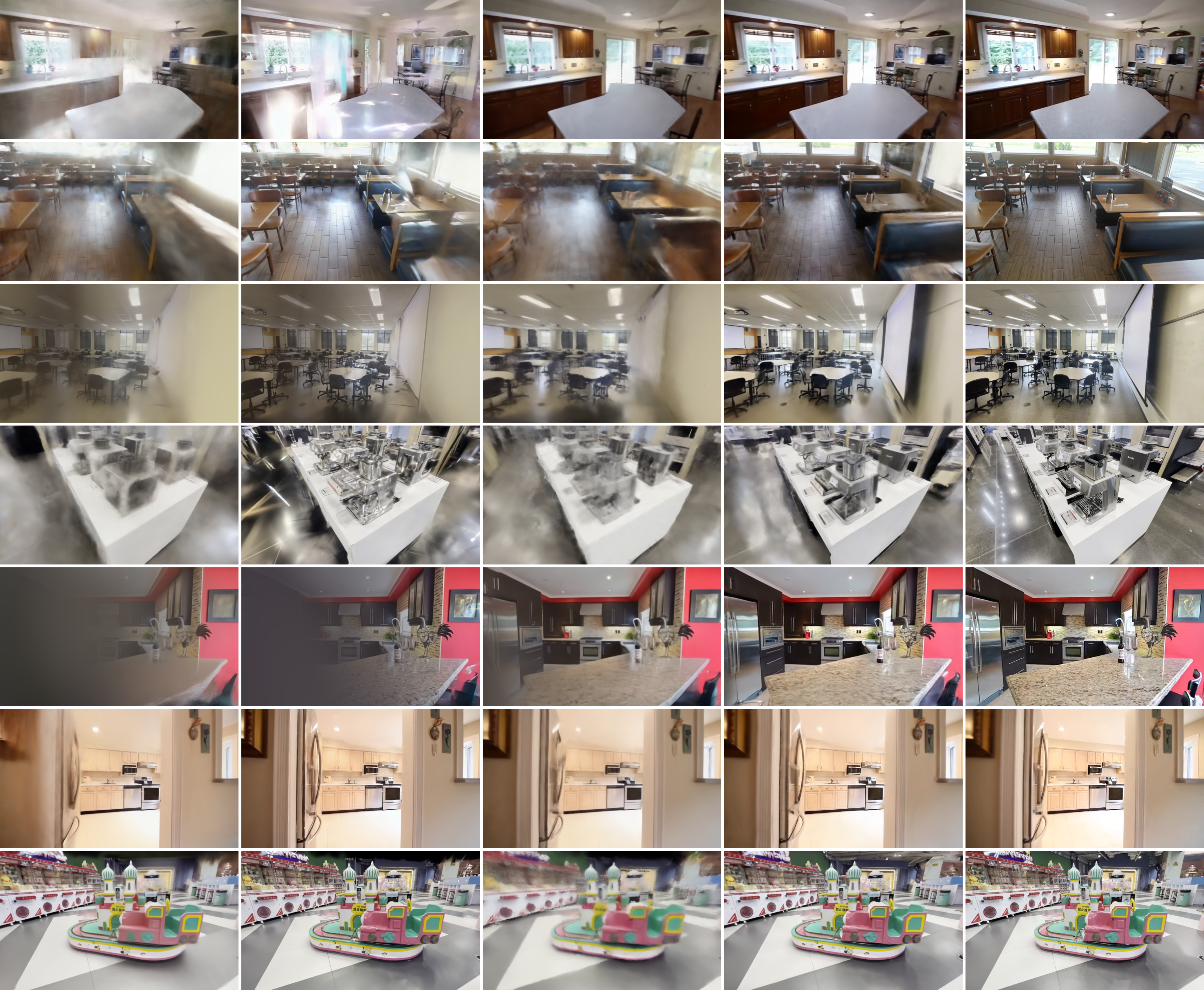}%
    } \\
    \hspace{0.8cm} \small GenFusion~\cite{Wu2025GenFusion} & 
    \hspace{1.2cm}\small Difix3D+~\cite{wu2025difix3d} & 
    \hspace{1.2cm}\small ExploreGS~\cite{Kim_2025_ExploreGS} &
    \hspace{0.7cm}\small \ours (Ours) & 
    \hspace{-0.1cm}\small Ground Truth \\

\end{tabular}  
\caption{\textbf{Additional Qualitative Comparison on 3D Reconstruction on DL3DV~\cite{ling2024dl3dv} and RE10K~\cite{re10k}.} \ours showcases sharper and cleaner appearance and geometry compared to baseline methods.}
    \label{fig:supp_recon}
\end{figure*}

\paragraph{Additional Qualitative Comparison.}
We provide additional novel-view refinement comparison in \cref{fig:supp_image_refine}. The top three rows correspond to the results on optimization-based 3DGS reconstruction and the bottom three feed-forward model reconstruction.
Compared to methods in both categories, our method (\ours) achieves sharper textures, more consistent geometry, and superior visual refinement, closely matching the ground-truth images. \cref{fig:supp_recon} presents the qualitative comparison on refined 3DGS reconstruction. \ours excels at challenging cases where in the original reconstruction part of the geometry is missing/wrong: while baseline methods smear out the artifacts and blur out those regions, \ours can faithfully refine/repaint the geometry, demonstrating superior photorealism and cross-view consistency.

\section{Limitations and Future Work}
\label{supp:limitations}
\paragraph{Limitations.}
\ours, although better than baselines, performs less effectively on videos with rapid motion or severe motion blur, where the VAE struggles to encode clean latent representations, leading to degraded generation quality. Our model operates on a context window of 81 frames. To process longer videos, we adopt a bidirectional sliding-window strategy, where the last frame of the current prediction is used as the first frame of the next window. While this approach maintains short-term temporal consistency across a few hundred frames, it does not fully guarantee long-term coherence or synchronization between independently generated segments.

\paragraph{Future Work.}
Future extensions could explore improved distillation techniques, such as self-forcing~\cite{huang2025selfforcing} and incorporate causal modeling with a rolling KV cache, to better support streaming or real-time applications. Another promising direction is to develop mechanisms that enable the video generative model to directly update the 3D Gaussian parameters online, allowing dynamic scene refinement without post-hoc optimization.

\end{document}